\documentclass[11pt,a4paper]{article}
\usepackage[hyperref]{acl2021}
\usepackage{times}
\usepackage{latexsym}

\usepackage{microtype}

\usepackage{amsmath,amsfonts,bm}

\def\eqref#1{equation~\ref{#1}}

\def\1{\bm{1}}

\def\rb{{\textnormal{b}}}

\def\rd{{\textnormal{d}}}

\def\rh{{\textnormal{h}}}

\def\ry{{\textnormal{y}}}

\def\rvx{{\mathbf{x}}}

\def\rmW{{\mathbf{W}}}

\def\vx{{\bm{x}}}

\DeclareMathAlphabet{\mathsfit}{\encodingdefault}{\sfdefault}{m}{sl}
\SetMathAlphabet{\mathsfit}{bold}{\encodingdefault}{\sfdefault}{bx}{n}

\def\gL{{\mathcal{L}}}

\usepackage{amsmath}
\usepackage{multirow}
\usepackage{amssymb}
\usepackage{mathrsfs}
\usepackage{graphics}
\usepackage{graphicx}
\usepackage{subfigure}
\usepackage[rightcaption]{sidecap}
\usepackage{booktabs}
\usepackage{multirow}
\usepackage{array}
\usepackage{bbm}
\usepackage{multicol}
\usepackage{blindtext}
\usepackage{algorithm}
\usepackage{algpseudocode}
\usepackage{tikz}
\usepackage{makecell}
\usepackage{xcolor,colortbl}
\usepackage{threeparttable}
\usepackage{array}
\usepackage{float}

\aclfinalcopy 
\def\aclpaperid{2931} 

\def\eg{\emph{e.g. }}

\title{Enhancing the Generalization for Intent Classification\\and Out-of-Domain Detection in SLU}
\author{Yilin Shen, 
  Yen-Chang Hsu,
  Avik Ray,
  Hongxia Jin \\
  Samsung Research America \\
  \small{\texttt{\{yilin.shen,yenchang.hsu,avik.r,hongxia.jin\}@samsung.com}}
}

\date{}

\graphicspath {{figure/}}

\usepackage{pifont}

\newcommand*{\Scale}[2][4]{\scalebox{#1}{$#2$}}%

\def\@fnsymbol#1{\ensuremath{\ifcase#1\or *\or \dagger\or \ddagger\or
   \mathsection\or \mathparagraph\or \|\or **\or \dagger\dagger
   \or \ddagger\ddagger \else\@ctrerr\fi}}
\newcommand{\ssymbol}[1]{^{\@fnsymbol{#1}}}

\begin{document}
\maketitle
\begin{abstract}
Intent classification is a major task in spoken language understanding (SLU).
Since most models are built with pre-collected in-domain (IND) training utterances, their ability to detect unsupported out-of-domain (OOD) utterances has a critical effect in practical use.
Recent works have shown that using extra data and labels can improve the OOD detection performance, yet it could be costly to collect such data.
This paper proposes to train a model with only IND data while supporting both IND intent classification and OOD detection.
Our method designs a novel domain-regularized module (DRM) to reduce the overconfident phenomenon of a vanilla classifier, achieving a better generalization in both cases.
Besides, DRM can be used as a drop-in replacement for the last layer in any neural network-based intent classifier, providing a low-cost strategy for a significant improvement.
The evaluation on four datasets shows that our method built on BERT and RoBERTa models achieves state-of-the-art performance against existing approaches and the strong baselines we created for the comparisons.
\end{abstract}

\section{Introduction}

Spoken language understanding (SLU) systems play a crucial role in ubiquitous artificially intelligent voice-enabled personal assistants (PA).
SLU needs to process a wide variety of user utterances and carry out user's intents, a.k.a. \emph{intent classification}.
Many deep neural network-based SLU models have recently been proposed and have demonstrated significant progress \citep{guo2014joint,liu2016attention,Zhang2016slu,Yu2018bimodel,Goo2018slotgated,Chen2019bertslu} in classification accuracy.
These models usually apply the closed-world assumption, in which the SLU model is trained with predefined domains, and the model expects to see the same data distribution during both training and testing.
However, such an assumption is not held in the practical use case of PA systems, where the system is used under a dynamic and open environment with personal expressions, new vocabulary, and unknown intents that are out of the design scope.

\begin{figure}[t]
	\centering
	\includegraphics[width=0.9\columnwidth]{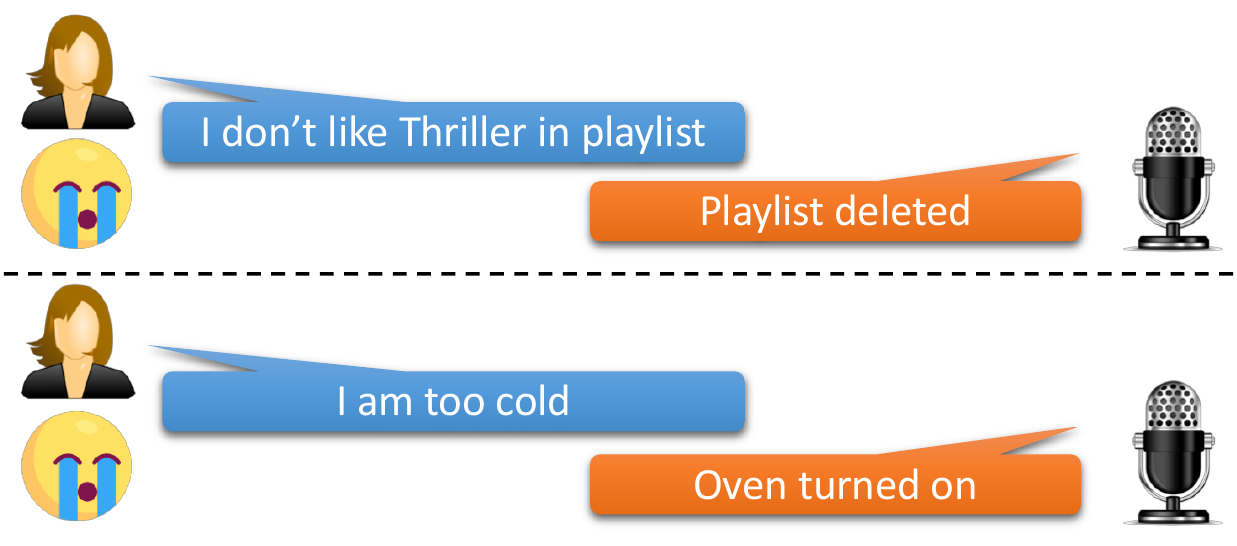}
	\caption{Failure Examples of Unsupported Skills in AI Voice Assistants. The user's utterances are out of the designed domains of the assistant.}
	\label{fig:ood_example}
\end{figure}

To address the challenges in open-world settings, previous works adopt varied strategies.  \citet{shen2018cruise,shen2019QA} use a cold-start algorithm to generate additional training data to cover a larger variety of utterances. This strategy relies on the software developers to pre-build all possible skills. \citet{shen2019skillbot,Shen2019teach} introduce a SkillBot that allows users to build up their own skills.
Recently, \citet{RayShenJin:18,ray2019fast,Shen2018interspeech,shen2019slot} enables an SLU model to incorporate user personalization over time.
However, the above approaches do not explicitly address unsupported user utterances/intents, leading to catastrophic failures illustrated in~\autoref{fig:ood_example}.
Thus, it is critically desirable for an SLU system to classify the supported intents (\textbf{in-domain (IND)}) and reject unsupported ones (\textbf{out-of-domain (OOD)}) correctly.

A straightforward solution is to collect OOD data and train a supervised binary classifier on both IND data and OOD data \citep{hendrycks2018OE}. However, collecting a representative set of OOD data could be impractical due to the infinite compositionality of language.
Arbitrarily selecting a subset could incur the selection bias, causing the learned model might not generalize to unseen OOD data.
\citet{Ryu2017ae,ryu2018ood} avoid learning with OOD data by using generative models (\textit{e.g.,} autoencoder and GAN) to capture the IND data distribution, then judge IND/OOD based on the reconstruction error or likelihood.
Recently, \citet{tan2019ood} utilizes a large training data to enable the meta-learning for OOD detection. \citet{zheng2019ood} generates pseudo OOD data to learn the OOD detector. 
The above-discussed approaches require additional data or training procedures beyond the intent classification task, introducing significant data collection effort or inference overhead. 

This paper proposes a strategy based on neural networks to use only IND utterances and their labels to learn both the intent classifier and OOD detector. Our strategy modifies the structure of the classifier, introducing an extra branch as a regularization target. We call the structure a \emph{Domain-Regularized Module} (DRM). This structure is probabilistically motivated and empirically leads to a better generalization in both intent classification and OOD detection. Our analysis focuses more on the latter task, finding that DRM not only outputs a class probability that is a better indicator for judging IND/OOD, but also leads to a feature representation with a less distribution overlap between IND and OOD data. More importantly, DRM is a simple drop-in replacement of the last linear layer, making it easy to plug into any off-the-shelf pre-trained models (\eg BERT \citep{devlin2018bert}) to fine-tune for a target task. The evaluation on four datasets shows that DRM can consistently improve upon previous state-of-the-art methods.

\section{Problem Definition \& Background}\label{sc:background}

\subsection{Problem Definition}

In the application of intent classification, a user utterance will be either an in-domain (IND) utterance (supported by the system) or an out-of-domain (OOD) utterance (not supported by the system). The classifier is expected to correctly (1) predict the intent of supported IND utterances; and (2) detect to reject the unsupported OOD utterances.

The task is formally defined below.
We are given a closed world IND training set $D_{IND}= \{\rvx, \ry\} = \{(\vx_i, y_i)\}^N_{i=1}$.
Each sample $(\vx_i, y_i)$, an utterance $\vx_i$ and its intent class label $y_i \in \{1 \ldots C\}$ for $C$ predefined in-domain classes, is drawn from a fixed but unknown IND distribution $P_{IND}(\rvx, \ry)$.
We aim to train an intent classifier model \emph{only} on IND training data $D_{IND}$ such that the model can perform:
(1) \textbf{Intent Classification:} classify the intent class label $y$ of an utterance $\vx$ if $\vx$ is drawn from the same distribution $P_{IND}$ as the training set $D_{IND}$;
(2) \textbf{OOD Detection:} detect an utterance $\vx$ to be an abnormal/unsupported sample if  $\vx$ is drawn from a different distribution $P_{OOD}$.

\subsection{Related Work}

\hspace*{5mm}\textbf{Intent Classification} 
is one of the major SLU components \citep{HafTurWri:03,WangDengAce:05,tur2011spoken}.
Various models have been proposed to encode the user utterance for intent classification, including RNN \citep{ravuri2015comparative, Zhang2016slu, liu2016attention,KimLeeStratos:17,Yu2018bimodel,Goo2018slotgated}, Recursive autoencoders \citep{kato2017utterance}, or enriched word embeddings \citep{kim2016intent}.
Recently, the BERT model \citep{devlin2018bert} was explored by \citep{Chen2019bertslu} for SLU.
Our work also leverages the representation learned in BERT.

\textbf{OOD Detection}
has been studied for many years \citep{hellman1970nearest}. \citet{tur2014detecting} explores its combination with intent classification by learning an SVM classifier on the IND data and randomly sampled OOD data. \citet{Ryu2017ae} detects OOD by using reconstruction criteria with an autoencoder. \citet{ryu2018ood} learns an intent classifier with GAN and uses the discriminator as the classifier for OOD detection. \citet{zheng2019ood} leverages extra unlabeled data to generate pseudo-OOD samples using GAN via auxiliary classifier regularization. \citet{tan2019ood} further incorporates the few-shot setting, learning the encoding of sentences with a prototypical network that is regularized with the OOD data outside a learning episode.
Other researchers developed methods in computer vision based on the rescaling of the predicted class probabilities (ODIN) \citep{liang2017enhancing} or building the Gaussian model with the features extracted from the hidden layers of neural networks (Mahalanobis) \citep{lee2018simple}.
Recently, \citep{hsu2020generalized} proposed Generalized-ODIN with decomposed confidence scores.
However, both approaches also heavily depend on the image input perturbation to achieve good performance.
Unfortunately, such perturbation cannot be applied to discrete utterance data in SLU.

\section{Our Method}

Our method is inspired by the decomposed confidence of Generalized-ODIN \citep{hsu2020generalized}, but we leverage the fact that the training data are all from IND to introduce an extra regularization. This regularization leads to a better generalization (lower classification error) on the intent classification. The improvement is in contrast to the original Generalized-ODIN, which has its classification error slightly increased. Since the improved generalization is likely due to a more generalizable feature representation, we leverage this observation, providing a modified Mahalanobis \citep{lee2018simple}, which we called L-Mahalanobis, for a transformer-based model to detect OOD data. In the following sections, we first describe the DRM and then elaborate on using the outputs of a DRM-equipped model to detect OOD data.

\subsection{Domain-Regularized Module (DRM)}

The motivation begins with introducing the domain variable $d$ ($d=1$ means IND, while $d=0$ means OOD) following the intuition in~\cite{hsu2020generalized}, then rewrite the posterior of class $y$ given $x$ with domain $d$ as follows:
\begin{eqnarray}\label{eq:dis_posterior}
	& \widehat{p}(\ry| \rd=1, \vx) &
	= \frac{\widehat{p}(\ry, \rd=1 |\vx)}{\widehat{p}(\rd=1|\vx)} \nonumber \\
	&&= \frac{\widehat{p}(\ry |\vx)}{\widehat{p}(\rd=1|\vx)} - \frac{\widehat{p}(\ry,  \rd=0|\vx)}{\widehat{p}(\rd=1|\vx)} \nonumber \\
	&&\approx \frac{\widehat{p}(\ry|\vx)}{\widehat{p}(\rd=1|\vx)}
\end{eqnarray}
where the last step holds since $\widehat{p}(\ry,  \rd=0|\vx)$ is close to 0 with the intrinsic conflict between IND classes $\ry$ and random variable $\rd=0$ for OOD.

\subsubsection{DRM Design}

\begin{figure}[t]
	\centering
	\includegraphics[width=1\columnwidth]{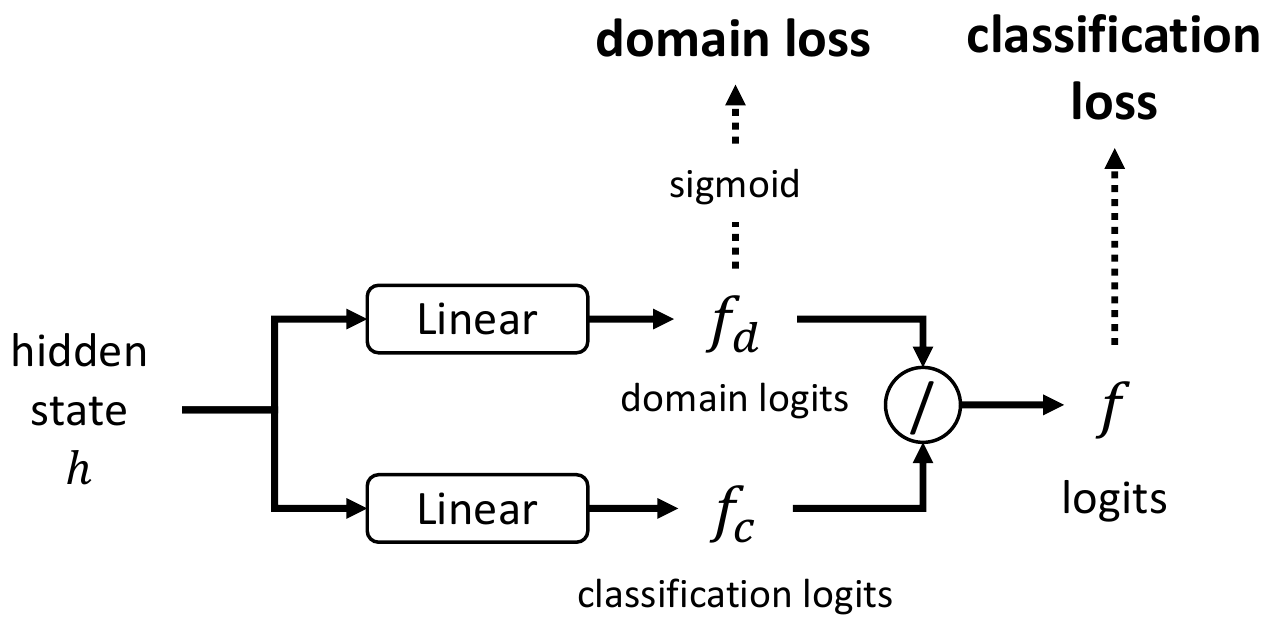}
	\caption{The DRM involves a domain component and a classification components for the IND classes.}
	\label{fig:disentanglement}
\end{figure}

Motivated by the above~\autoref{eq:dis_posterior}, we design the DRM to mitigate overconfidence by decomposing the final logits $f$ into two branches.
\autoref{fig:disentanglement} illustrates the architecture.

\noindent\textbf{Domain Logits $f_d$} models $\widehat{p}(\rd=1|\vx)$ before normalization.
It projects from hidden state $\rh$ to a scalar w.r.t. $\rd$:
\begin{equation}\label{eq:fd}
f_d = \rmW_d \rh + \rb_d
\end{equation}
where $\rmW_d \in \mathcal{R}^{|h| \times 1}$.
Since $\widehat{p}(\rd=1|\vx)$ is a probability between 0 and 1, Section \ref{sc:training} will describe the training details of domain loss via the sigmoid function.

\noindent\textbf{Classification Logits $f_c$} models the probability posterior $\widehat{p}(\ry |\vx)$ before normalization.
It follows the conventional linear projection from hidden state $\rh$ to the number of classes:
\begin{equation}
f_c = \rmW_c \rh + \rb_c
\end{equation}
where $\rmW_d \in \mathcal{R}^{|h| \times C}$ with $C$ classes.

At the end, we obtain the final logits $f$ to represent $\widehat{p}(\ry| \rd=1, \vx)$ by putting $f_d$ and $f_c$ together following the dividend-divisor structure of ~\autoref{eq:dis_posterior}:
\begin{equation}\label{eq:dis_logits}
f = f_c / f_d
\end{equation}
where each element of $f_c$ is divided by the same scalar $f_d$.

\subsubsection{DRM Training}\label{sc:training}

We propose two training loss functions to train a model with DRM.
The first training loss aims to minimize a cross-entropy between the predicted intent class and ground truth IND class labels. 
\begin{equation}
\gL_{\textsl{classification}} \triangleq - \sum_{i=1}^{C} \ry^i \log \widehat{p}(f)^i
\end{equation}
where $\widehat{p}(f)$ is the softmax of logits $f$:
\begin{equation*}
\widehat{p}(f) = \textsl{softmax} (f)
\end{equation*}

The second training loss aims to ensure that the domain component $f_d$ is close to 1 since all utterances in the training set are IND.
\begin{equation} \label{eq:domain_loss}
\gL_{\textsl{domain}} \triangleq (1 - \textsl{sigmoid} (f_d))^2
\end{equation}
We first restrict $f_d$ between 0 and 1 by using \textsl{sigmoid} activation function.
Then, this loss function encourages $\textsl{sigmoid} (f_d)$ close to 1 for training on IND utterances.
In order to avoid $f_d$ to be very large values and affect the training convergence, we further apply clamp function on $f_d$ before it feeds to~\autoref{eq:dis_logits}:
\begin{equation*}
f_d =
\begin{cases}
f_d & \mbox{if } -\delta < f_d < \delta \\
\delta & \mbox{if } f_d <= -\delta \mbox{ or } f_d >= \delta \\
\end{cases}
\end{equation*}
Thus, we sum them up to optimize the model:
\begin{equation}
\gL = \gL_{\textsl{classification}} + \gL_{\textsl{domain}}
\end{equation}

\emph{
Remarks:
It is important to note that the design of $\gL_{\textsl{domain}}$ is to introduce extra regularization to mitigate the overconfidence in standard posterior probability $\widehat{p}(f)$. $\textsl{sigmoid} (f_d)$ is not used to directly predict if an utterance is IND or OOD.
}

\setlength{\tabcolsep}{1.4em}
\begin{table*}[t]\scriptsize
  \centering
  \caption{SLU Benchmark and In-House Dataset Statistics}
    \begin{tabular}{c|c|cccc}
    \toprule
    Dataset & Domain & \#Intents & \#Train & \#Dev & \#Test \\
    \midrule
    \midrule
    \multirow{2}{*}{CLINC \citep{larson2019ooddataset}} & various domains in voice assistants & 150   & 15,000 & 3,000 & 4,500 \\
          & other out-of-scope domains & -     & 100   & 100   & 1,000 \\
    \midrule
    ATIS \citep{Hemphill1990ATIS} & airline travel information domain & 18    & 4,478 & 500   & 893 \\
    \midrule
    Snips \citep{Alice2018snips} & music, book, and weather domains & 7     & 13,084 & 700   & 700 \\
    \midrule
    Movie (in-house) & movie QA domain & 38    & 39,558 & 4,897 & 4,926 \\
    \bottomrule
    \end{tabular}%
  \label{tab:datasets}%
\end{table*}%

\subsection{IND Intent Classification Method}

Following~\autoref{eq:dis_posterior} and our DRM design, it is straightforward to use the confidence score of $\textsl{softmax}(f)$ to predict the IND intent class.

\subsection{OOD Detection Methods}

There are two types of strategies to utilize the outputs of a classifier to perform OOD detection.
One is based on the confidence which is computed from logits, the other is based on the features. In the below, we describe how to compute different OOD scores with our DRM.

\subsubsection{Confidence-based Methods}

Recent works \citep{liang2017enhancing} has shown that the softmax outputs provide a good scoring for detecting OOD data.
In our DRM model, we use the \emph{decomposed softmax outputs} for the score.
The logits $f_c$ w.r.t. the true posterior distribution in open-world can be combined with varied approaches:

\textbf{DRM Confidence Score:}
\begin{equation}
\textsl{Conf}_{DRM} = \textsl{softmax} (f_c)
\end{equation}

\textbf{DRM ODIN Confidence Score:}
\begin{equation}
\textsl{ODIN}_{DRM} = \textsl{softmax} (f_c/T)
\end{equation}
with large $T=1000$ \citep{liang2017enhancing}.

\textbf{DRM Entropy Confidence Score:}
\begin{equation}
\textsl{ENT}_{DRM} = \textsl{Entropy} [\textsl{softmax} (f_c)]
\end{equation}

The OOD utterances have low $\textsl{Conf}_{DRM}$, $\textsl{ODIN}_{DRM}$ scores and high $\textsl{ENT}_{DRM}$ score.

\subsubsection{Feature-based Method}

While our DRM confidence already outperforms many existing methods (later shown in experiments), we further design the feature-based Mahalanobis distance score, inspired by the recent work \citep{lee2018simple} for detecting OOD images. 

We first recap the approach in \citep{lee2018simple} which consists of two parts: Mahalanobis distance calculation and input preprocessing.
Mahalanobis distance score models the class conditional Gaussian distributions w.r.t. Gaussian discriminant analysis based on both low- and upper-level features of the deep classifier models.
The score on layer $\ell$ is computed as follows:
\begin{equation*}
\Scale[0.9]{
S^{\ell}_{Maha}(\vx)=\max_i -(f^{\ell}(\vx) - \mu^{\ell}_c)^T \Sigma_{\ell}^{-1} (f^{\ell}(\vx) - \mu^{\ell}_c)
}
\end{equation*}
where $f^{\ell}(\vx)$ represents the output features at the $\ell^{th}$-layer of neural networks;
$\mu_i$ and $\Sigma$ are the class mean representation and the covariance matrix.
Thus, the overall score is their summation:
\begin{equation*}
S_{Maha}(\vx)=\sum_{\ell} S_{Maha}(f^{\ell}(\vx))
\end{equation*}
In addition, the input preprocessing adds a small controlled noise to the test samples to enhance the performance.

Although Mahalanobis distance score can be applied only to the last feature layer without input preprocessing $S^{\textsl{last}}_{Maha}(\vx)$, the analysis (Table 2 in \citep{lee2018simple}) shows that either input preprocessing or multi-layer scoring mechanism is required to achieve decent OOD detection performance.
Unfortunately, neither of the above two mechanisms is applicable in the intent classifier for SLU.
First, unlike image data, noise injection into discrete natural language utterances has been shown not to perform well.
Second, in most cutting-edge intent classifier models, low- and upper-level network layers are quite different.
The direct application of multi-layer Mahalanobis distance leads to much worse OOD detection performance.

\begin{figure}[t]
	\centering
	\includegraphics[width=1\columnwidth]{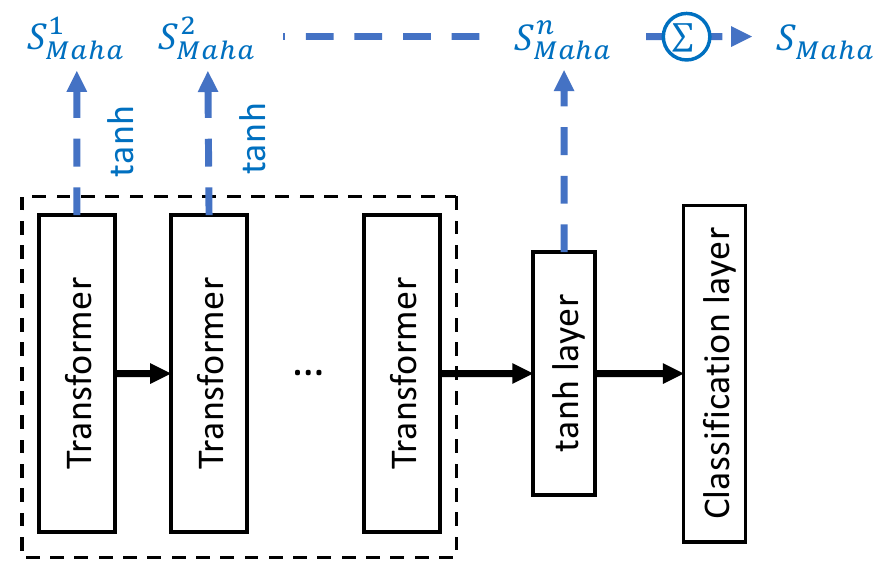}
	\caption{Multi-layer Mahalanobis Score Design for BERT-based Classifier Model}
	\label{fig:maha}
\end{figure}

Since BERT-based models showed significant performance improvement for intent classification in SLU \citep{Chen2019bertslu}, we focus on designing the multi-layer Mahalanobis score for BERT-based classifier models.
In existing BERT-based text classification models, such as BERT, RoBERTa, DistilBERT, ALBERT, etc., there are different designs between the last transformer layer and the classification layer.
~\autoref{fig:maha} shows our generic design of Mahalanobis score computation (blue) for various BERT-based models.

Our design is based on our extensive experiments and understanding of the common insights in different BERT-based models. Specifically, we use the features from different layers between the last transformer layer and the classification layer. We empirically found that the nonlinear tanh layer plays an important role.
Thus, to map the features of each transformer layer and last layer into the same semantic space, we pass the features of each layer through tanh function and sum them up to compute our Mahalanobis score:
\begin{eqnarray}
    S_{L-Maha}(\vx)=S_{Maha}(f^{n}(\vx)) \nonumber \\
    + \sum_{1 \le \ell < n} S_{Maha}(\textsl{tanh}(f^{\ell}(\vx)))
\end{eqnarray}
where $f^{\ell}$ and $f^{n}$ are the features of each layer $\ell$ and last layer $n$ in a BERT-based intent classifier model.
We refer to our proposed approach as \textbf{L-Mahalanobis}.

\section{Experimental Evaluation}

\subsection{Datasets}

We evaluate our proposed approach on three benchmark SLU datasets and one in-house SLU dataset.
~\autoref{tab:datasets} provides an overview of all datasets.
Among all these datasets, the recently released CLINC dataset serves as a benchmark for OOD detection in SLU.
For the other three datasets, we treat them mutually OOD due to non-overlapping domains.

We crowdsourced the in-house Movie dataset containing common questions that users may ask regarding movies.
This dataset mainly consists of queries a user may ask in the movie domain. The dataset consists of $38$ different intents (e.g. rating information, genre information, award information, show trailer) and $20$ slots or entities (e.g., director, award, release year). This dataset was collected using crowdsourcing as follows. At first, some example template queries were generated by linguistic experts for each intent, along with intent and slot descriptions. Next, a generation crowdsourcing job was launched where a crowd worker was assigned a random intent, a combination of entities, and few slots generally associated with the intent. To better understand the intent and slots, the worker was asked to review the intent and slot descriptions, and example template utterances. The first task of the worker was to provide $3$ different queries corresponding to the given intent, which also contains the provided entities. The second task of the worker was to provide additional entities corresponding to the same slot type. A subsequent validation crowdsourcing job was launched where these crowdsourced queries were rated by validation workers in terms of their accuracy with the provided intent and entities. Each query was rated by $5$ different validation workers, and the final validated dataset contains a subset of crowdsourced queries with high accuracy score and high inter-rater agreement.

\subsection{Implementation and Training Details}

We implemented our method using PyTorch on top of the Hugging Face transformer library \citep{Wolf2019HuggingFacesTS}.
We follow the hyperparameters in the original models.
For the only hyperparameter $\delta$, we experimented only on CLINC dataset from 2.2 to 4 with uniform interval 0.2 (we try 10 values of $\delta$) based on $\textsl{sigmoid}(2.2)\approx 0.9$ and $\textsl{sigmoid}(4)\approx 0.982$.
We used $\delta=3$ which gives the best performance in our experiment for all datasets.
We train each model with 3 epochs using 4 NVIDIA Tesla V100 GPUs (16GB) for each training.
We conducted experiments on two transformer-based models, BERT \citep{devlin2018bert} and RoBERTa \citep{Liu2019roberta}.

\emph{
Remarks: All experiments only use IND data for both training and validation.
We use the same hyperparameters in all datasets and validate the generalizability of our method.
}

\subsection{Baselines}

\subsubsection{IND Intent Classification Baselines}
We consider the strongest baseline \textbf{BERT-Linear} (the last layer is linear) fine-tuned on the pre-trained BERT-based models \citep{Chen2019bertslu}.

\subsubsection{OOD Detection Baselines}
We consider the existing OOD detection methods: 

\textbf{ConGAN \citep{ryu2018ood}:}
a GAN-based model based on given sentence representations to generate OOD features with additional feature matching loss.
OOD utterances are expected to have low discriminator confidence scores.

\textbf{Autoencoder (AE) \citep{Ryu2017ae}:}
first uses an LSTM based classifier model to train sentence representations;
then train an autoencoder on the above sentence embeddings.
OOD utterances are expected to have high reconstruction error.

\textbf{ODIN \citep{liang2017enhancing}:} 
we only use the temperature scaling on logits.
OOD utterances are expected to have a low scaled confidence score.

\textbf{Generalized-ODIN (G-ODIN) \citep{hsu2020generalized}:} 
we fine-tune on pre-trained BERT models with replaced last layer and only use the decomposed confidence.
We evaluate all three variations proposed in the paper $h^I$, $h^E$ and $h^C$ and report the best one.
OOD utterances are expected to have low scaled confidence score.

\textbf{Mahalanobis \citep{lee2018simple}:}
we only use the feature of BERT's last layer to compute Mahalanobis distance score.
OOD utterances are expected to have a low scaled confidence score.

For ConGAN and AE, we evaluate the model in the original paper as well as customized BERT-based backbone models as strong baselines.
Specifically, we customize \textbf{En-ConGAN} and \textbf{En-AE} as follows:
\textbf{En-ConGAN} uses BERT sentence representation as input;
\textbf{En-AE} applies a BERT classifier model to train the sentence representation and then use them to further train an autoencoder.
\emph{Thus, En-ConGAN and En-AE are not existing baselines.}

Note that ERAEPOG \citep{zheng2019ood} and O-Proto \citep{tan2019ood} are not comparable since they require additional unlabeled data and labels.
We only put the ERAEPOG results on CLINC dataset (from the original paper) for reference.

\setlength{\tabcolsep}{0.65em}
\begin{table*}[thbp!]\scriptsize
  \centering
  \caption{Comprehensive OOD Detection Results on CLINC Dataset (CLINC Train/OOD)}
    \begin{tabular}{c|c|c|cccccc}
    \toprule
    \multirow{2}[2]{*}{Model} & \multirow{2}[2]{*}{Last Layer} & \multirow{2}[2]{*}{OOD Method} & \multicolumn{6}{c}{OOD Evaluation} \\
          &       &       & EER($\downarrow$) & FPR95($\downarrow$) & Detection Error($\downarrow$) & AUROC($\uparrow$) & AUPR In($\uparrow$) & AUPR Out($\uparrow$) \\
    \midrule
    \midrule
    ConGAN & -   & -   & 78.90$\ssymbol{4}$ & 94.40$\ssymbol{4}$ & 52.04$\ssymbol{4}$ & 52.22$\ssymbol{4}$ & 82.79$\ssymbol{4}$ & 23.54$\ssymbol{4}$ \\
    AE  & -   & -   & 18.13$\ssymbol{4}$ & 58.50$\ssymbol{4}$ & 23.94$\ssymbol{4}$ & 87.78$\ssymbol{4}$ & 96.98$\ssymbol{4}$ & 54.12$\ssymbol{4}$ \\
    \textcolor[rgb]{ .651,  .651,  .651}{ERAEPOG} & \textcolor[rgb]{ .651,  .651,  .651}{-} & \textcolor[rgb]{ .651,  .651,  .651}{-} & \textcolor[rgb]{ .651,  .651,  .651}{12.04}$\ssymbol{4}$ & \textcolor[rgb]{ .651,  .651,  .651}{23.70}$\ssymbol{4}$ & \textcolor[rgb]{ .651,  .651,  .651}{11.67}$\ssymbol{4}$ & \textcolor[rgb]{ .651,  .651,  .651}{95.83}$\ssymbol{4}$ & \textcolor[rgb]{ .651,  .651,  .651}{99.05}$\ssymbol{2}$ & \textcolor[rgb]{ .651,  .651,  .651}{83.98}$\ssymbol{4}$ \\
    
    \midrule
    \midrule
     
     \multirow{14}{*}{BERT} & \multicolumn{2}{c|}{En-ConGAN} & 75.20$\ssymbol{4}$ & 98.72$\ssymbol{4}$ & 49.95$\ssymbol{4}$ & 22.36$\ssymbol{4}$ & 69.86$\ssymbol{4}$ & 11.27$\ssymbol{4}$ \\
     & \multicolumn{2}{c|}{En-AE} & 8.70$\ssymbol{4}$ & 13.03$\ssymbol{4}$ & 8.47$\ssymbol{4}$ & 96.12$\ssymbol{4}$ & 98.89$\ssymbol{4}$ & 88.38$\ssymbol{4}$ \\
     & \multicolumn{2}{c|}{ODIN} & 9.01$\ssymbol{4}$ & 16.52$\ssymbol{4}$ & 8.66$\ssymbol{4}$ & 96.24$\ssymbol{4}$ & 98.73$\ssymbol{4}$ & 87.34$\ssymbol{4}$ \\
     & \multicolumn{2}{c|}{G-ODIN} & 8.91$\ssymbol{4}$ & 12.99$\ssymbol{4}$ & 8.40$\ssymbol{4}$ & 95.81$\ssymbol{4}$ & 98.75$\ssymbol{2}$ & 88.81$\ssymbol{4}$ \\
      \cmidrule{2-9}
      & \multirow{4}{*}{Linear} & Confidence & 11.31$\ssymbol{4}$ & 21.98$\ssymbol{4}$ & 11.00$\ssymbol{4}$ & 94.96$\ssymbol{4}$ & 98.52$\ssymbol{4}$ & 84.59$\ssymbol{4}$ \\
         &    & Entropy & 10.33$\ssymbol{4}$ & 17.99$\ssymbol{4}$ & 10.10$\ssymbol{4}$ & 95.65$\ssymbol{4}$ & 98.73$\ssymbol{4}$ & 87.20$\ssymbol{4}$ \\
         &    & Mahalanobis & 8.31$\ssymbol{4}$  & 12.68$\ssymbol{4}$ & 8.02$\ssymbol{4}$  & 96.90$\ssymbol{4}$ & 99.14$\ssymbol{2}$ & 88.19$\ssymbol{4}$ \\
         &    & L-Mahalanobis* & 7.21  & 10.18 & 6.92  & 97.52 & 99.41 & 89.37 \\
      \cmidrule{2-9}
         & \multicolumn{1}{c|}{\multirow{4}{*}{DRM*}} & Confidence* & 8.50 & 12.85 & 7.85 & 96.34 & 98.95 & 87.51 \\
         &    & Entropy* & 8.31 & 12.53 & 8.14 & 96.67 & 99.01 & 89.68 \\
         &    & Mahalanobis* & 7.01  & 10.88 & 6.88  & 97.43 & 99.37 & 90.36 \\
         &    & L-Mahalanobis* & \textbf{6.70} & \textbf{10.12} & \textbf{6.62} & \textbf{97.77} & \textbf{99.46} & \textbf{91.55} \\
    
    \midrule
    \midrule
  
       \multirow{14}{*}{RoBERTa} & \multicolumn{2}{c|}{En-ConGAN} & 80.26$\ssymbol{4}$ & 99.34$\ssymbol{4}$ & 49.95$\ssymbol{4}$ & 15.20$\ssymbol{4}$ & 66.64$\ssymbol{4}$ & 10.58$\ssymbol{4}$ \\
     & \multicolumn{2}{c|}{En-AE} & 8.56$\ssymbol{4}$ & 12.38$\ssymbol{4}$ & 8.29$\ssymbol{4}$ & 96.82$\ssymbol{4}$ & 99.08$\ssymbol{2}$ & 90.06$\ssymbol{4}$ \\
     & \multicolumn{2}{c|}{ODIN} & 9.11$\ssymbol{4}$ & 15.12$\ssymbol{4}$ & 8.68$\ssymbol{4}$ & 96.11$\ssymbol{4}$ & 98.84$\ssymbol{4}$ & 88.72$\ssymbol{4}$ \\
     & \multicolumn{2}{c|}{G-ODIN} & 8.85$\ssymbol{4}$ & 12.26$\ssymbol{4}$ & 8.53$\ssymbol{4}$ & 96.74$\ssymbol{4}$ & 99.12$\ssymbol{2}$ & 89.95$\ssymbol{4}$ \\
      \cmidrule{2-9}
      & \multirow{4}{*}{Linear} & Confidence & 10.81$\ssymbol{4}$ & 22.35$\ssymbol{4}$ & 10.38$\ssymbol{4}$ & 95.23$\ssymbol{4}$ & 98.58$\ssymbol{4}$ & 86.46$\ssymbol{4}$ \\
         &    & Entropy & 9.31$\ssymbol{4}$ & 14.81$\ssymbol{4}$ & 8.93$\ssymbol{4}$ & 95.89$\ssymbol{4}$ & 98.73$\ssymbol{4}$ & 88.70$\ssymbol{4}$ \\
         &    & Mahalanobis & 8.40$\ssymbol{4}$  & 11.82$\ssymbol{4}$  & 8.13$\ssymbol{4}$  & 96.92$\ssymbol{4}$ & 99.06$\ssymbol{4}$ & 90.37$\ssymbol{4}$ \\
         &    & L-Mahalanobis* & 6.90  & 9.53  & 6.71  & 97.94 & 99.50 & 92.47 \\
      \cmidrule{2-9}
         & \multicolumn{1}{c|}{\multirow{4}{*}{DRM*}} & Confidence* & 8.35 & 11.76 & 8.02 & 97.10 & 99.25 & 90.46 \\
         &    & Entropy* & 8.29 & 11.51 & 7.86 & 97.17 & 99.27 & 90.69 \\
         &    & Mahalanobis* & 6.31  & 7.80  & 6.13  & 98.07 & 99.53 & 92.86 \\
         &    & L-Mahalanobis* & \textbf{6.11} & \textbf{7.63} & \textbf{5.98} & \textbf{98.16} & \textbf{99.56} & \textbf{92.96} \\

    \bottomrule
    \end{tabular}
    \begin{tablenotes}
     \item[1] Our best method (DRM+L-Mahalanobis) is significantly better than each baseline model (without *) with $\textsl{p-value}<0.01$ (marked by $\mathsection$) and $\textsl{p-value}<0.05$ (marked by $\dagger$) using t-test.
     All methods with * are our proposed methods.
    \end{tablenotes}
  \label{tab:clinc_results}
\end{table*}

\subsection{Evaluation Metrics}

\subsubsection{IND Intent Classification Metrics}
We evaluate IND performance using the \textbf{classification accuracy} metric as in literature \citep{liu2016attention,Yu2018bimodel,Chen2019bertslu}.

\subsubsection{OOD Detection Metrics}
we follow the evaluation metrics in literature \citep{ryu2018ood} and \citep{liang2017enhancing,lee2018simple}. 
Let TP, TN, FP, and FN denote true positive, true negative, false positive, and false negative.
We use the following OOD evaluation metrics:

\textbf{EER (lower is better)}: (Equal Error Rate)
measures the error rate when false positive rate (FPR) is equal to the false negative rate (FNR).
Here, FPR=FP/(FP+TN) and FNR=FN/(TP+FN).

\textbf{FPR95 (lower is better)}: (False Positive Rate (FPR) at 95\% True Positive Rate (TPR))
can be interpreted as the probability that an OOD utterance is misclassified as IND when the true positive rate (TPR) is as high as 95\%.
Here, TPR=TP/(TP+FN).

\textbf{Detection Error (lower is better)}:
measures the misclassification probability when TPR is 95\%.
Detection error is defined as follows:
\begin{eqnarray*}
    \min_\delta \{ P_{IND}(s\le \delta) p(\vx \in P_{IND}) \\
    + P_{OOD}(s > \delta) p(\vx \in P_{OOD}) \}
\end{eqnarray*}
where $s$ is a confidence score.
We follow the same assumption that both IND and OOD examples have an equal probability of appearing in the testing set.

\textbf{AUROC (higher is better)}: (Area under the Receiver Operating Characteristic Curve)
The ROC curve is a graph plotting TPR against the FPR=FP/(FP+TN) by varying a threshold.

\textbf{AUPR (higher is better)}: (Area under the Precision-Recall Curve (AUPR))
The PR curve is a graph plotting the
precision against recall by varying a threshold.
Here, precision=TP/(TP+FP) and recall=TP/(TP+FN).
AUPR-IN and AUPR-OUT is AUPR where IND and OOD distribution samples are specified as positive, respectively.

Note that EER, detection error, AUROC, and AUPR are threshold-independent metrics.

\subsubsection{Statistical Significance}
We also evaluate the statistical significance between all baselines and our best result (DRM + L-Mahalanobis) on all the above metrics.
We train each model 10 times with different PyTorch random seeds.
We report the average results and t-test statistical significance results.

\subsection{Results}

\subsubsection{IND Classification Results}

\autoref{tab:ind_results} reports the IND intent classification results on each dataset finetuned using BERT and RoBERTa pre-trained models.
It is interesting to observe that all DRM combined models consistently achieve better classification accuracy with up to 0.8\% improvement (reproduced "No joint" row in Table 3 in \citep{Chen2019bertslu} on Snips dataset).
This is because the domain loss forces $\textsl{sigmoid} (f_d)$ close to 1 and therefore also slightly mitigates its impact to IND classification.
Thus, the true posterior distribution of IND data is also modeled more precisely.
For both BERT and RoBERTa backbones, DRM models are significantly better than conventional BERT-linear classification models with $\textsl{p-value}<0.05$.

\setlength{\tabcolsep}{0.6em}
\begin{table}[t]\scriptsize
  \centering
  \caption{IND Intent Classification Results}
    \begin{tabular}{cc|cccc}
    \toprule
    \multirow{2}{*}{Model} & \multirow{2}{*}{Last Layer} & \multicolumn{4}{c}{Datasets} \\
          &       & CLINC & ATIS  & Snips & Movie \\
    \midrule
    \midrule
    \multirow{2}{*}{BERT} & Linear & 96.19$\ssymbol{2}$ & 97.76$\ssymbol{2}$ & 97.97$\ssymbol{2}$ & 97.26$\ssymbol{2}$ \\
          & DRM*  & \textbf{96.66} & \textbf{98.21} & \textbf{98.23} & \textbf{97.87} \\
    \midrule
    \multirow{2}{*}{RoBERTa} & Linear & 96.82$\ssymbol{2}$ & 97.64$\ssymbol{2}$ & 98.07$\ssymbol{2}$ & 98.07$\ssymbol{2}$ \\
          & DRM*  & \textbf{97.15} & \textbf{98.31} & \textbf{98.87} & \textbf{98.63} \\
    \bottomrule
    \end{tabular}%
    \begin{tablenotes}
     \item [1] Our DRM methods (marked by *) are significantly better than baseline model on all datasets with $\textsl{p-value}<0.05$ (marked by $\dagger$) using t-test.
   \end{tablenotes}
  \label{tab:ind_results}%
\end{table}%

\subsubsection{OOD Detection Results}

\setlength{\tabcolsep}{0.2em}
\begin{table*}[thbp!]\scriptsize
  \centering
  \caption{OOD Detection Results on Snips/ATIS/Movie Datasets (RoBERTa Model Finetuning)}
    \begin{tabular}{c|c|c|c|c|c|c}
    \toprule
    \multirow{2}[2]{*}{OOD Method} & \multicolumn{6}{c}{OOD Evaluation} \\
          & EER($\downarrow$) & FPR95($\downarrow$) & Detection Error($\downarrow$) & AUROC($\uparrow$) & AUPR In($\uparrow$) & AUPR Out($\uparrow$) \\
    \midrule
    \midrule
    
    \multicolumn{7}{c}{IND dataset: Snips;    OOD Datasets: CLINC\_OOD/ATIS/Movie} \\
    \midrule
    En-ConGAN  & 54.50$\ssymbol{4}$/63.05$\ssymbol{4}$/54.22$\ssymbol{4}$  & 99.16$\ssymbol{4}$/99.87$\ssymbol{4}$/99.10$\ssymbol{4}$  & 42.61$\ssymbol{4}$/49.10$\ssymbol{4}$/37.32$\ssymbol{4}$  & 39.03$\ssymbol{4}$/30.88$\ssymbol{4}$/45.64$\ssymbol{4}$  & 37.15$\ssymbol{4}$/34.47$\ssymbol{4}$/30.03$\ssymbol{4}$  & 51.23$\ssymbol{4}$/45.70$\ssymbol{4}$/52.59$\ssymbol{4}$ \\
    Confidence  & 9.91$\ssymbol{4}$/17.83$\ssymbol{4}$/22.22$\ssymbol{4}$  & 14.94$\ssymbol{4}$/47.43$\ssymbol{4}$/51.85$\ssymbol{4}$  & 9.18$\ssymbol{4}$/11.17$\ssymbol{4}$/19.34$\ssymbol{4}$  & 96.09$\ssymbol{4}$/92.03$\ssymbol{4}$/87.44$\ssymbol{4}$  & 94.78$\ssymbol{4}$/92.65$\ssymbol{4}$/97.67$\ssymbol{4}$  & 97.21$\ssymbol{4}$/92.29$\ssymbol{4}$/55.16$\ssymbol{4}$ \\
    Entropy  & 10.21$\ssymbol{4}$/18.05$\ssymbol{4}$/23.15$\ssymbol{4}$  & 14.54$\ssymbol{4}$/45.04$\ssymbol{4}$/52.68$\ssymbol{4}$  & 9.25$\ssymbol{4}$/10.77$\ssymbol{4}$/19.58$\ssymbol{4}$  & 96.32$\ssymbol{4}$/92.44$\ssymbol{4}$/87.12$\ssymbol{4}$  & 94.90$\ssymbol{4}$/92.94$\ssymbol{4}$/97.60$\ssymbol{4}$  & 97.53$\ssymbol{4}$/92.99$\ssymbol{4}$/52.27$\ssymbol{4}$ \\
    ODIN  & 10.01$\ssymbol{4}$/16.93$\ssymbol{4}$/23.15$\ssymbol{4}$  & 14.22$\ssymbol{4}$/39.04$\ssymbol{4}$/58.33$\ssymbol{4}$  & 9.43$\ssymbol{4}$/9.64$\ssymbol{4}$/23.01$\ssymbol{4}$  & 96.46$\ssymbol{4}$/93.81$\ssymbol{4}$/83.58$\ssymbol{4}$  & 94.59$\ssymbol{4}$/93.99$\ssymbol{4}$/96.63$\ssymbol{4}$  & 97.75$\ssymbol{4}$/94.53$\ssymbol{4}$/47.36$\ssymbol{4}$ \\
    G-ODIN  & 9.65$\ssymbol{4}$/15.16$\ssymbol{4}$/22.02$\ssymbol{4}$  & 13.31$\ssymbol{4}$/37.86$\ssymbol{4}$/55.67$\ssymbol{4}$  & 8.32$\ssymbol{4}$/8.55$\ssymbol{4}$/21.82$\ssymbol{4}$  & 97.21$\ssymbol{4}$/94.73$\ssymbol{4}$/85.60$\ssymbol{4}$  & 95.70$\ssymbol{4}$/95.04$\ssymbol{4}$/97.73$\ssymbol{4}$  & 98.02$\ssymbol{4}$/95.44$\ssymbol{4}$/50.38$\ssymbol{4}$ \\
    En-AE  & 4.40$\ssymbol{4}$/4.37$\ssymbol{4}$/3.59$\ssymbol{4}$  & 4.18$\ssymbol{4}$/3.59$\ssymbol{4}$/3.08$\ssymbol{2}$  & 4.25$\ssymbol{4}$/4.00$\ssymbol{4}$/3.64$\ssymbol{2}$  & 98.56$\ssymbol{4}$/98.12$\ssymbol{4}$/88.96$\ssymbol{4}$  & 97.41$\ssymbol{4}$/98.92$\ssymbol{2}$/94.39$\ssymbol{4}$  & 98.12$\ssymbol{4}$/95.34$\ssymbol{4}$/86.84$\ssymbol{4}$ \\
    Maha & 3.90$\ssymbol{4}$/1.81/11.11$\ssymbol{4}$ & 2.66$\ssymbol{4}$/2.23$\ssymbol{4}$/5.58$\ssymbol{4}$ & 3.47$\ssymbol{4}$/1.36$\ssymbol{2}$/10.21$\ssymbol{4}$ & 98.79$\ssymbol{2}$/99.74$\ssymbol{2}$/95.61$\ssymbol{4}$ & 97.73$\ssymbol{4}$/99.75$\ssymbol{2}$/99.22$\ssymbol{4}$ & 99.21$\ssymbol{4}$/99.77$\ssymbol{2}$/76.61$\ssymbol{4}$ \\
    DRM+L-Maha* & \textbf{3.00/1.79/2.78} & \textbf{1.95/0.00/2.78} & \textbf{2.63/1.16/3.16} & \textbf{98.90/99.79/98.53} & \textbf{98.15/99.79/99.76} & \textbf{99.24/99.80/87.02} \\
    
    \midrule
    \midrule
    
    \multicolumn{7}{c}{IND dataset: ATIS;    OOD Datasets: CLINC\_OOD/Snips/Movie} \\
    \midrule
    En-ConGAN  & 21.60$\ssymbol{4}$/19.74$\ssymbol{4}$/23.28$\ssymbol{4}$  & 81.52$\ssymbol{4}$/86.33$\ssymbol{4}$/93.77$\ssymbol{4}$  & 15.51$\ssymbol{4}$/15.54$\ssymbol{4}$/16.03$\ssymbol{4}$  & 82.34$\ssymbol{4}$/81.79$\ssymbol{4}$/79.32$\ssymbol{4}$  & 84.52$\ssymbol{4}$/89.35$\ssymbol{4}$/58.36$\ssymbol{4}$  & 72.74$\ssymbol{4}$/60.20$\ssymbol{4}$/89.14$\ssymbol{4}$ \\
    Confidence  & 10.21$\ssymbol{4}$/8.52$\ssymbol{4}$/10.19$\ssymbol{4}$  & 20.50$\ssymbol{4}$/12.92$\ssymbol{4}$/17.59$\ssymbol{4}$  & 9.28$\ssymbol{4}$/8.36$\ssymbol{4}$/9.33$\ssymbol{4}$  & 96.99$\ssymbol{4}$/97.84$\ssymbol{4}$/96.62$\ssymbol{4}$  & 97.19$\ssymbol{4}$/98.57$\ssymbol{4}$/99.56$\ssymbol{2}$  & 97.04$\ssymbol{4}$/96.99$\ssymbol{4}$/84.27$\ssymbol{4}$ \\
    Entropy  & 9.91$\ssymbol{4}$/8.84$\ssymbol{4}$/10.12$\ssymbol{4}$  & 21.67$\ssymbol{4}$/13.75$\ssymbol{4}$/17.59$\ssymbol{4}$  & 9.11$\ssymbol{4}$/8.16$\ssymbol{4}$/9.38$\ssymbol{4}$  & 97.06$\ssymbol{4}$/97.93$\ssymbol{4}$/96.68$\ssymbol{4}$  & 97.25$\ssymbol{4}$/98.62$\ssymbol{4}$/99.57$\ssymbol{2}$  & 97.11$\ssymbol{4}$/97.14$\ssymbol{4}$/85.02$\ssymbol{4}$ \\
    ODIN  & 9.11$\ssymbol{4}$/8.36$\ssymbol{4}$/10.08$\ssymbol{4}$  & 21.32$\ssymbol{4}$/14.39$\ssymbol{4}$/18.52$\ssymbol{4}$  & 7.50$\ssymbol{4}$/6.15$\ssymbol{4}$/9.37$\ssymbol{4}$  & 97.16$\ssymbol{4}$/98.00$\ssymbol{4}$/96.73$\ssymbol{4}$  & 97.39$\ssymbol{4}$/98.68$\ssymbol{4}$/99.58$\ssymbol{2}$  & 97.16$\ssymbol{4}$/97.18$\ssymbol{4}$/84.88$\ssymbol{4}$ \\
    G-ODIN  & 8.75$\ssymbol{4}$/8.01$\ssymbol{4}$/9.97$\ssymbol{4}$  & 20.87$\ssymbol{4}$/13.44$\ssymbol{4}$/17.76$\ssymbol{4}$  & 7.31$\ssymbol{4}$/6.02$\ssymbol{4}$/8.98$\ssymbol{4}$  & 97.27$\ssymbol{4}$/98.11$\ssymbol{4}$/96.85$\ssymbol{4}$  & 97.46$\ssymbol{4}$/98.76$\ssymbol{4}$/99.59$\ssymbol{2}$  & 97.28$\ssymbol{4}$/97.32$\ssymbol{4}$/85.90$\ssymbol{4}$ \\
    En-AE & 4.00$\ssymbol{4}$/\textbf{2.09}/3.69$\ssymbol{4}$ & 2.20$\ssymbol{2}$/\textbf{0.00}/0.35$\ssymbol{2}$ & 3.45$\ssymbol{2}$/\textbf{1.33}/1.97$\ssymbol{2}$ & 99.41$\ssymbol{2}$/\textbf{99.83}/99.63$\ssymbol{4}$ & 99.43$\ssymbol{2}$/\textbf{99.89}/98.72$\ssymbol{4}$ & 99.43$\ssymbol{2}$/\textbf{99.74}/97.93$\ssymbol{4}$ \\
    Maha & 4.00$\ssymbol{4}$/3.85$\ssymbol{4}$/6.48$\ssymbol{4}$ & 12.13$\ssymbol{4}$/8.06$\ssymbol{4}$/11.64$\ssymbol{4}$ & 3.76$\ssymbol{4}$/2.94$\ssymbol{4}$/5.04$\ssymbol{4}$ & 99.18$\ssymbol{4}$/99.47$\ssymbol{4}$/98.72$\ssymbol{2}$ & 98.78$\ssymbol{4}$/99.45$\ssymbol{4}$/99.71$\ssymbol{4}$ & 99.46/99.49/95.45$\ssymbol{4}$ \\
    DRM+L-Maha* & \textbf{2.70/2.09/1.85} & \textbf{1.30}/0.32/\textbf{0.00} & \textbf{2.55}/2.01/\textbf{1.23} & \textbf{99.48}/99.70/\textbf{99.78} & \textbf{99.51}/99.82/\textbf{99.97} & \textbf{99.47}/99.50/\textbf{98.22} \\
    
    \midrule
    \midrule
    
    \multicolumn{7}{c}{IND dataset: Movie;    OOD Datasets: CLINC\_OOD/ATIS/Snips} \\
    \midrule
    En-ConGAN  & 45.90$\ssymbol{4}$/15.12$\ssymbol{4}$/41.09$\ssymbol{4}$  & 44.05$\ssymbol{4}$/14.35$\ssymbol{4}$/39.59$\ssymbol{4}$  & 22.95$\ssymbol{4}$/7.56$\ssymbol{4}$/20.55$\ssymbol{4}$  & 43.85$\ssymbol{4}$/57.44$\ssymbol{4}$/45.78$\ssymbol{4}$  & 85.21$\ssymbol{4}$/88.23$\ssymbol{4}$/90.40$\ssymbol{4}$  & 14.68$\ssymbol{4}$/17.56$\ssymbol{4}$/10.09$\ssymbol{4}$ \\
    Confidence  & 19.22$\ssymbol{4}$/16.70$\ssymbol{4}$/18.81$\ssymbol{4}$  & 36.81$\ssymbol{4}$/47.94$\ssymbol{4}$/47.52$\ssymbol{4}$  & 18.51$\ssymbol{4}$/15.15$\ssymbol{4}$/18.53$\ssymbol{4}$  & 91.65$\ssymbol{4}$/91.99$\ssymbol{4}$/90.53$\ssymbol{4}$  & 98.11$\ssymbol{4}$/98.50$\ssymbol{4}$/98.68$\ssymbol{4}$  & 76.78$\ssymbol{4}$/67.58$\ssymbol{4}$/59.63$\ssymbol{4}$ \\
    Entropy  & 19.12$\ssymbol{4}$/17.26$\ssymbol{4}$/19.13$\ssymbol{4}$  & 34.64$\ssymbol{4}$/44.24$\ssymbol{4}$/44.80$\ssymbol{4}$  & 18.25$\ssymbol{4}$/16.12$\ssymbol{4}$/18.87$\ssymbol{4}$  & 91.79$\ssymbol{4}$/92.14$\ssymbol{4}$/90.72$\ssymbol{4}$  & 98.11$\ssymbol{4}$/98.50$\ssymbol{4}$/98.69$\ssymbol{4}$  & 78.66$\ssymbol{4}$/70.87$\ssymbol{4}$/63.96$\ssymbol{4}$ \\
    ODIN  & 19.42$\ssymbol{4}$/18.95$\ssymbol{4}$/19.94$\ssymbol{4}$  & 34.43$\ssymbol{4}$/39.91$\ssymbol{4}$/39.38$\ssymbol{4}$  & 18.24$\ssymbol{4}$/18.38$\ssymbol{4}$/19.33$\ssymbol{4}$  & 91.34$\ssymbol{4}$/91.40$\ssymbol{4}$/90.03$\ssymbol{4}$  & 97.96$\ssymbol{4}$/98.29$\ssymbol{4}$/98.53$\ssymbol{4}$  & 78.56$\ssymbol{4}$/71.62$\ssymbol{4}$/65.18$\ssymbol{4}$ \\
    G-ODIN  & 18.61$\ssymbol{4}$/18.23$\ssymbol{4}$/19.25$\ssymbol{4}$  & 34.19$\ssymbol{4}$/36.42$\ssymbol{4}$/37.03$\ssymbol{4}$  & 18.15$\ssymbol{4}$/17.27$\ssymbol{4}$/18.91$\ssymbol{4}$  & 91.86$\ssymbol{4}$/91.97$\ssymbol{4}$/90.63$\ssymbol{4}$  & 98.21$\ssymbol{4}$/98.34$\ssymbol{4}$/98.70$\ssymbol{4}$  & 78.98$\ssymbol{4}$/72.07$\ssymbol{4}$/66.79$\ssymbol{4}$ \\
    En-AE  & 13.70$\ssymbol{4}$/7.28$\ssymbol{4}$/16.05$\ssymbol{4}$  & 43.42$\ssymbol{4}$/16.05$\ssymbol{4}$/32.29$\ssymbol{4}$  & 11.00$\ssymbol{4}$/4.46$\ssymbol{4}$/11.87$\ssymbol{4}$  & 94.57$\ssymbol{4}$/93.56$\ssymbol{4}$/92.23$\ssymbol{4}$  & 98.91$\ssymbol{4}$/99.58$\ssymbol{2}$/99.01$\ssymbol{2}$  & 77.12$\ssymbol{4}$/76.13$\ssymbol{4}$/68.75$\ssymbol{4}$ \\
    Maha & 3.90$\ssymbol{4}$/3.41$\ssymbol{2}$/6.11$\ssymbol{4}$ & 6.02$\ssymbol{4}$/2.35$\ssymbol{4}$/15.40$\ssymbol{4}$ & 3.72$\ssymbol{4}$/3.02$\ssymbol{4}$/6.02$\ssymbol{4}$ & 99.37$\ssymbol{4}$/99.43$\ssymbol{4}$/98.63$\ssymbol{4}$ & 99.81$\ssymbol{2}$/99.89/99.82$\ssymbol{2}$ & 97.82$\ssymbol{2}$/97.27$\ssymbol{4}$/91.44$\ssymbol{4}$ \\
    DRM+L-Maha* & \textbf{3.70/3.36/4.66} & \textbf{2.56/1.01/4.34} & \textbf{3.61/2.85/4.58} & \textbf{99.48/99.53/99.06} & \textbf{99.89/99.92/99.88} & \textbf{97.90/97.38/93.85} \\
    
    \bottomrule
    \end{tabular}%
    \begin{tablenotes}
      \item[1] In each OOD method for an IND dataset, "/" separates the results for different OOD datasets.
      \item[1] Our method (*) is significantly better than baseline models with $\textsl{p-value}<0.01$ (marked by $\mathsection$) and $\textsl{p-value}<0.05$ (marked by $\dagger$) using t-test in most cases.
    \end{tablenotes}
  \label{tab:ood_results}%
\end{table*}%

\textbf{Results on CLINC Dataset:}
\autoref{tab:clinc_results} reports the OOD detection results on CLINC dataset.
This result covers all existing work and our enhanced baselines.
We focus on analyzing the contribution by each of our proposed techniques, DRM and L-Mahalanobis.
The first three rows report the performance of existing approaches based on the original designs in their papers (ERAEPOG in grey uses additional unlabeled data).
Unfortunately, we observe that their performance is even worse than the simple confidence-based approach via BERT finetuning baseline (row 5).
Thus, we mainly focus on comparing our method with strong baselines with BERT and RoBERTa models.

For a given OOD detection method, we find that their combinations with DRM consistently perform better than those with standard models.
The improvement is at least 1-2\% for all metrics against our enhanced baselines.
Among all OOD detection approaches, our proposed L-Mahalanobis OOD detection approach achieves the best performance for both linear and DRM combined BERT and RoBERTa models.
It is not surprising to observe that our DRM method combined with a better pre-trained RoBERTa model achieves larger OOD detection performance improvement.
Note that our customized En-AE performs much better than most other methods since we incorporated the enhanced reconstruction capability with pre-trained BERT models.
However, En-AE cannot utilize all BERT layers as our proposed L-Mahalanobis method, resulting in worse performance.

In addition, DRM+L-Mahalanobis models are significantly better than existing methods and enhanced baselines with $\textsl{p-value}<0.01$ on most metrics for both BERT and RoBERTa backbones.

\noindent\textbf{Ablation Study on CLINC Dataset:}
We analyze how our two novel components, DRM model and L-Mahalanobis, impact the performance.

The rows with ``DRM" in ``Last Layer" column of~\autoref{tab:clinc_results} show the performance of DRM model.
As one can see, for all OOD methods, DRM consistently performs better than the conventional ``Linear" last layer.
Specifically, the DRM and Confidence combo also outperforms its closest baseline G-ODIN.
This validates the effectiveness of our disentangled logits design in DRM based on the mathematical analysis of overconfidence.
It also shows that our new domain loss can indeed enhance the model awareness that all training data is IND.

The rows with ``L-Mahalanobis" in ``OOD Method" column of~\autoref{tab:clinc_results} outperform other OOD methods with the same model and last layer.
Compared with its closest baseline Mahalanobis, the better performance of L-Mahalanobis validates the usefulness of all layers' features in various models.

\begin{figure*}[thbp!]
\centering
  \subfigure[Conventional Confidence Score]{
    \includegraphics[width=0.3\textwidth]{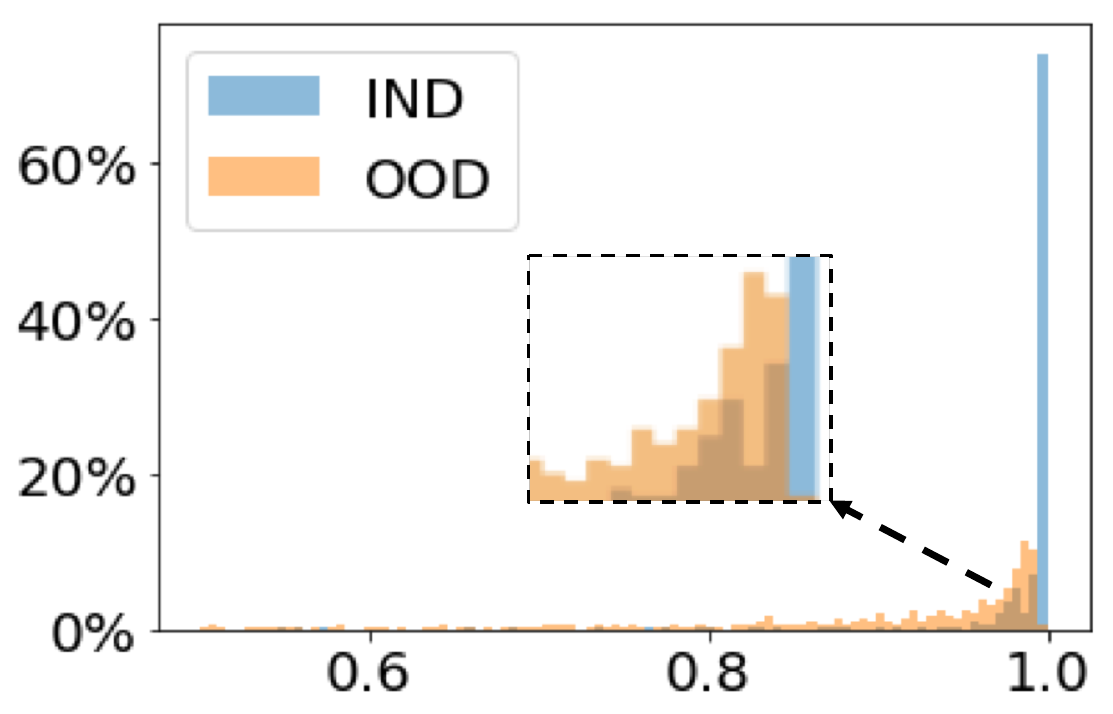}\label{fig:histogram_original}
  }
  \subfigure[DRM Confidence Score]{
    \includegraphics[width=0.3\textwidth]{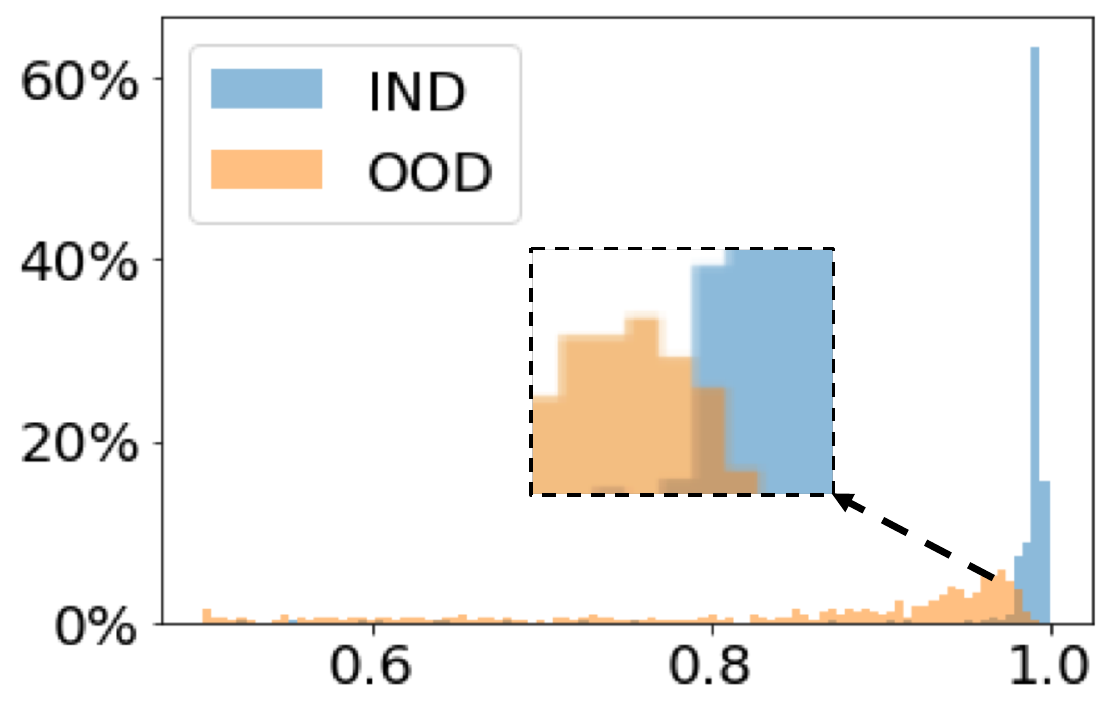}\label{fig:histogram_DRM}
  }
  \subfigure[DRM L-Mahalanobis Score]{
    \includegraphics[width=0.32\textwidth]{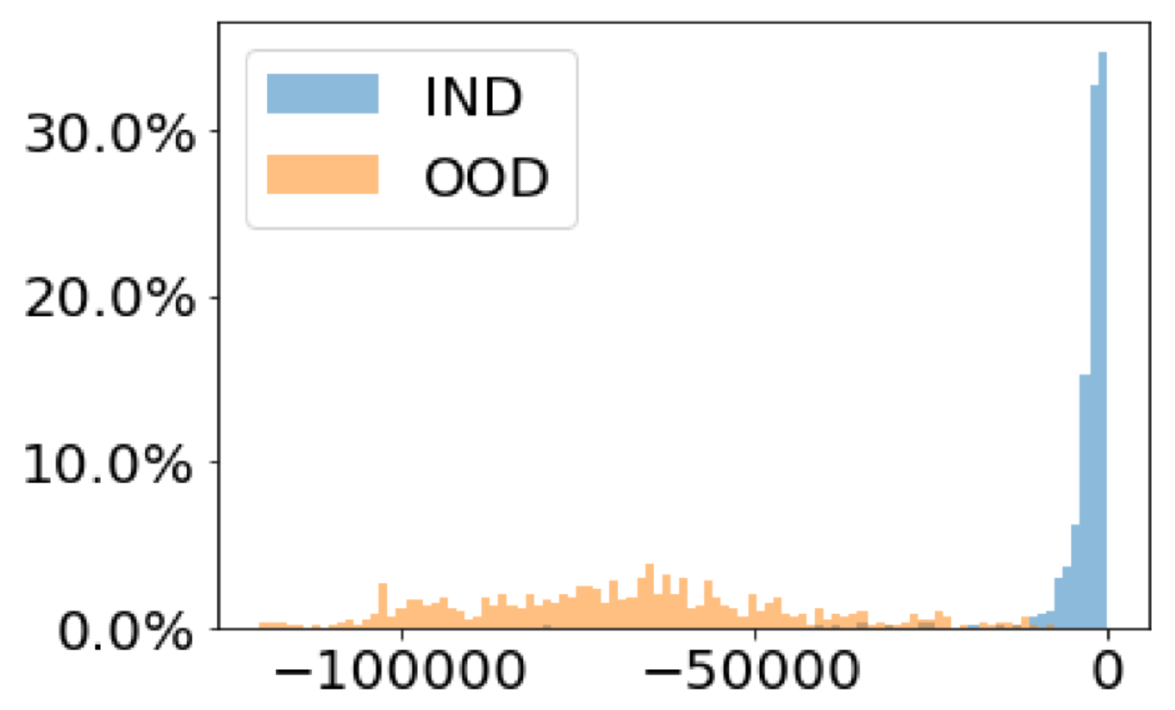}\label{fig:histogram_maha}
  }
\caption{Histogram of Detection Scores using Various Methods (Snips IND, ATIS OOD) (We choose this IND/OOD combination to provide the best visualization for analysis)}
\label{fig:histogram}
\end{figure*}

\noindent\textbf{Results on ATIS/Snips/Movie Datasets:}
Since our strong baselines on pre-trained RoBERTa model showed better results on CLINC, we next evaluate other results finetuned on RoBERTa model.
When taking each dataset as IND, we use the other two mutually exclusive datasets and CLINC\_OOD as OOD datasets for evaluating OOD detection performance.
As one can see in~\autoref{tab:ood_results}, our method outperforms other approaches on both Snips and movie IND datasets.
For ATIS IND dataset, En-AE for Snips OOD dataset achieves almost perfect performance.
This is because ATIS and Snips are almost completely non-overlapping and ATIS is well designed with carefully selected varieties and entities in the airline travel domain.
When taking Snip as IND and ATIS as OOD, it is interesting to see that our method achieves better performance than En-AE.
This is because that Snips contains a large number of entities such that the reconstruction error will be lower and become less separable than that in ATIS OOD utterances.

For both Snips and Movie IND datasets, DRM+L-Mahalanobis are significantly better than baseline methods with $\textsl{p-value}<0.01$ in most cases for all OOD datasets.
For ATIS IND dataset, DRM+L-Mahalanobis shows similar behavior except En-AE since it is easier to train an autoencoder model for ATIS IND dataset due to its carefully collected clean training utterances.

\subsection{Qualitative Analysis}

We provide a quantitative analysis by visualizing our two methods, DRM and L-Mahalanobis.

\subsubsection{Detection Score Distribution}
\autoref{fig:histogram} plots the histograms of detection scores for OOD and IND data.
Compared with~\autoref{fig:histogram_original}, DRM significantly reduces the overlap between OOD and IND in~\autoref{fig:histogram_DRM}.
L-Mahalanobis utilizes features from all layers to further reduce the overlap in~\autoref{fig:histogram_maha}.
Moreover, the score distributions from left to right in~\autoref{fig:histogram}, imply that a larger entropy of all score reflects a better uncertainty modeling.

\subsubsection{Feature Distribution Visualization}
\begin{figure}[h]
\centering
  \subfigure[Conventional RoBERTa]{
    \includegraphics[width=0.46\columnwidth]{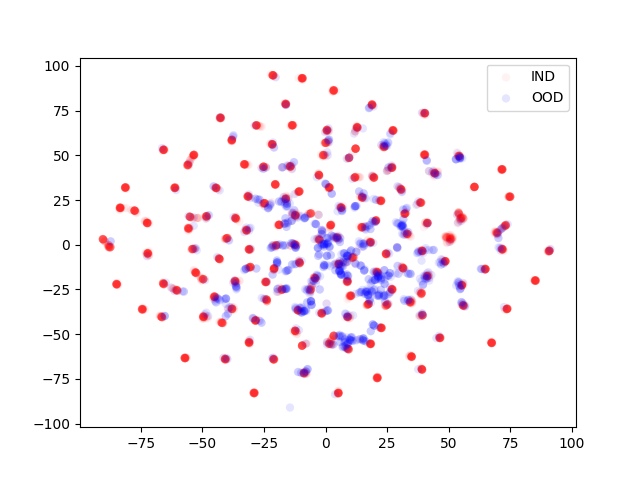}\label{fig:tsne_roberta_original}
  }
  \subfigure[DRM RoBERTa]{
    \includegraphics[width=0.46\columnwidth]{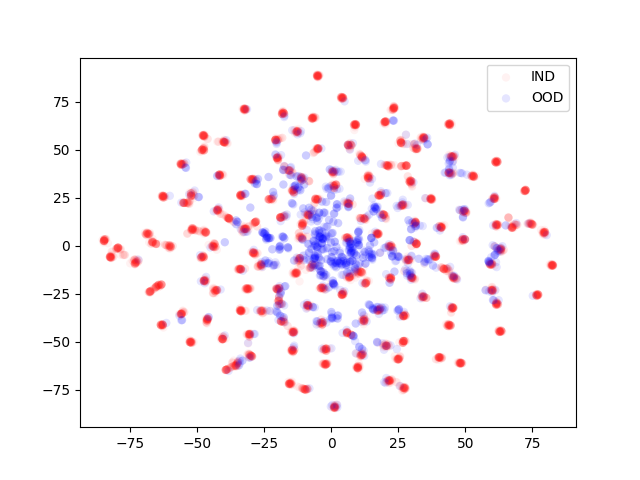}\label{fig:tsne_dis_roberta}
  }
\caption{t-SNE Visualization of Utterance Representations on CLINC Dataset (Red: IND, Blue: OOD)}
\label{fig:tsne_roberta}
\end{figure}

\autoref{fig:tsne_roberta} visualizes the utterance representations learned with or without DRM.
The red IND data are tightly clustered within classes (totally 150 CLINC IND classes), while the blue OOD data spread arbitrarily. 
As one can see, the blue dots in~\autoref{fig:tsne_dis_roberta} have less overlap with red dots, indicating the DRM helps to learn the utterance representation to better disentangle IND and OOD data.

\section {Conclusion}

This paper proposes using only IND utterances to conduct intent classification and OOD detection for SLU in an open-world setting.
The proposed DRM has a structure of two branches to avoid overconfidence and achieves a better generalization. The evaluation shows that our method can achieve state-of-the-art performance on various SLU benchmark and in-house datasets for both IND intent classification and OOD detection.
In addition, thanks to the generic of our DRM design and with the recent extensive use of BERT on different data modalities, our work can contribute to improving both in-domain classification robustness and out-of-domain detection robustness for various classification models such as image classification, sound classification, vision-language classifications.

\bibliographystyle{acl_natbib}
\bibliography{anthology,ref}

\begin{thebibliography}{37}
\expandafter\ifx\csname natexlab\endcsname\relax\def\natexlab#1{#1}\fi

\bibitem[{Chen et~al.(2019)Chen, Zhuo, and Wang}]{Chen2019bertslu}
Qian Chen, Zhu Zhuo, and Wen Wang. 2019.
\newblock {BERT} for joint intent classification and slot filling.
\newblock \emph{CoRR}, abs/1902.10909.

\bibitem[{Coucke et~al.(2018)Coucke, Saade, Ball, Bluche, Caulier, Leroy,
  Doumouro, Gisselbrecht, Caltagirone, Lavril, Primet, and
  Dureau}]{Alice2018snips}
Alice Coucke, Alaa Saade, Adrien Ball, Th{\'{e}}odore Bluche, Alexandre
  Caulier, David Leroy, Cl{\'{e}}ment Doumouro, Thibault Gisselbrecht,
  Francesco Caltagirone, Thibaut Lavril, Ma{\"{e}}l Primet, and Joseph Dureau.
  2018.
\newblock Snips voice platform: an embedded spoken language understanding
  system for private-by-design voice interfaces.
\newblock \emph{CoRR}, abs/1805.10190.

\bibitem[{Devlin et~al.(2019)Devlin, Chang, Lee, and
  Toutanova}]{devlin2018bert}
Jacob Devlin, Ming-Wei Chang, Kenton Lee, and Kristina Toutanova. 2019.
\newblock {BERT}: Pre-training of deep bidirectional transformers for language
  understanding.
\newblock In \emph{NAACL-HLT}, pages 4171--4186.

\bibitem[{Goo et~al.(2018)Goo, Gao, Hsu, Huo, Chen, Hsu, and
  Chen}]{Goo2018slotgated}
Chih-Wen Goo, Guang Gao, Yun-Kai Hsu, Chih-Li Huo, Tsung-Chieh Chen, Keng-Wei
  Hsu, and Yun-Nung Chen. 2018.
\newblock Slot-gated modeling for joint slot filling and intent prediction.
\newblock In \emph{NAACL-HLT}, pages 753--757.

\bibitem[{Guo et~al.(2014)Guo, Tur, Yih, and Zweig}]{guo2014joint}
Daniel Guo, Gokhan Tur, Wen-tau Yih, and Geoffrey Zweig. 2014.
\newblock Joint semantic utterance classification and slot filling with
  recursive neural networks.
\newblock In \emph{SLT}, pages 554--559.

\bibitem[{Haffner et~al.(2003)Haffner, Tur, and Wright}]{HafTurWri:03}
Patrick Haffner, Gokhan Tur, and Jerry~H Wright. 2003.
\newblock Optimizing svms for complex call classification.
\newblock In \emph{ICASSP}, volume~1.

\bibitem[{Hellman(1970)}]{hellman1970nearest}
Martin~E Hellman. 1970.
\newblock The nearest neighbor classification rule with a reject option.
\newblock \emph{IEEE Transactions on Systems Science and Cybernetics},
  6(3):179--185.

\bibitem[{Hemphill et~al.(1990)Hemphill, Godfrey, Doddington
  et~al.}]{Hemphill1990ATIS}
Charles~T Hemphill, John~J Godfrey, George~R Doddington, et~al. 1990.
\newblock The atis spoken language systems pilot corpus.
\newblock In \emph{Proceedings of the DARPA speech and natural language
  workshop}, pages 96--101.

\bibitem[{Hendrycks et~al.(2018)Hendrycks, Mazeika, and
  Dietterich}]{hendrycks2018OE}
Dan Hendrycks, Mantas Mazeika, and Thomas~G Dietterich. 2018.
\newblock Deep anomaly detection with outlier exposure.
\newblock In \emph{ICLR}.

\bibitem[{Hsu et~al.(2020)Hsu, Shen, Jin, and Kira}]{hsu2020generalized}
Yen{-}Chang Hsu, Yilin Shen, Hongxia Jin, and Zsolt Kira. 2020.
\newblock Generalized {ODIN:} detecting out-of-distribution image without
  learning from out-of-distribution data.
\newblock In \emph{CVPR}, pages 10948--10957.

\bibitem[{Kato et~al.(2017)Kato, Nagai, Noda, Sumitomo, Wu, and
  Yamamoto}]{kato2017utterance}
Tsuneo Kato, Atsushi Nagai, Naoki Noda, Ryosuke Sumitomo, Jianming Wu, and
  Seiichi Yamamoto. 2017.
\newblock Utterance intent classification of a spoken dialogue system with
  efficiently untied recursive autoencoders.
\newblock In \emph{SIGDIAL}, pages 60--64.

\bibitem[{Kim et~al.(2016)Kim, Tur, Celikyilmaz, Cao, and Wang}]{kim2016intent}
Joo-Kyung Kim, Gokhan Tur, Asli Celikyilmaz, Bin Cao, and Ye-Yi Wang. 2016.
\newblock Intent detection using semantically enriched word embeddings.
\newblock In \emph{SLT}, pages 414--419.

\bibitem[{Kim et~al.(2017)Kim, Lee, and Stratos}]{KimLeeStratos:17}
Young{-}Bum Kim, Sungjin Lee, and Karl Stratos. 2017.
\newblock {ONENET:} joint domain, intent, slot prediction for spoken language
  understanding.
\newblock In \emph{2017 {IEEE} Automatic Speech Recognition and Understanding
  Workshop}, pages 547--553.

\bibitem[{Larson et~al.(2019)Larson, Mahendran, Peper, Clarke, Lee, Hill,
  Kummerfeld, Leach, Laurenzano, Tang, and mars}]{larson2019ooddataset}
Stefan Larson, Anish Mahendran, Joseph~J. Peper, Christopher Clarke, Andrew
  Lee, Parker Hill, Jonathan~K. Kummerfeld, Kevin Leach, Michael~A. Laurenzano,
  Lingjia Tang, and Jason mars. 2019.
\newblock An evaluation dataset for intent classification and out-of-scope
  prediction.
\newblock In \emph{EMNLP-IJCNLP}, pages 1311--1316.

\bibitem[{Lee et~al.(2018)Lee, Lee, Lee, and Shin}]{lee2018simple}
Kimin Lee, Kibok Lee, Honglak Lee, and Jinwoo Shin. 2018.
\newblock A simple unified framework for detecting out-of-distribution samples
  and adversarial attacks.
\newblock In \emph{NeurIPS}, pages 7167--7177.

\bibitem[{Liang et~al.(2017)Liang, Li, and Srikant}]{liang2017enhancing}
Shiyu Liang, Yixuan Li, and R~Srikant. 2017.
\newblock Enhancing the reliability of out-of-distribution image detection in
  neural networks.
\newblock In \emph{ICLR}.

\bibitem[{Liu and Lane(2016)}]{liu2016attention}
Bing Liu and Ian Lane. 2016.
\newblock Attention-based recurrent neural network models for joint intent
  detection and slot filling.
\newblock In \emph{INTERSPEECH}, pages 685--689.

\bibitem[{Liu et~al.(2019)Liu, Ott, Goyal, Du, Joshi, Chen, Levy, Lewis,
  Zettlemoyer, and Stoyanov}]{Liu2019roberta}
Yinhan Liu, Myle Ott, Naman Goyal, Jingfei Du, Mandar Joshi, Danqi Chen, Omer
  Levy, Mike Lewis, Luke Zettlemoyer, and Veselin Stoyanov. 2019.
\newblock Roberta: {A} robustly optimized {BERT} pretraining approach.
\newblock \emph{CoRR}, abs/1907.11692.

\bibitem[{Ravuri and Stoicke(2015)}]{ravuri2015comparative}
Suman Ravuri and Andreas Stoicke. 2015.
\newblock A comparative study of neural network models for lexical intent
  classification.
\newblock In \emph{ASRU}, pages 368--374.

\bibitem[{Ray et~al.(2018)Ray, Shen, and Jin}]{RayShenJin:18}
Avik Ray, Yilin Shen, and Hongxia Jin. 2018.
\newblock Learning out-of-vocabulary words in intelligent personal agents.
\newblock In \emph{IJCAI}, pages 4309--4315.

\bibitem[{Ray et~al.(2019)Ray, Shen, and Jin}]{ray2019fast}
Avik Ray, Yilin Shen, and Hongxia Jin. 2019.
\newblock Fast domain adaptation of semantic parsers via paraphrase attention.
\newblock In \emph{DeepLo@EMNLP-IJCNLP}, pages 94--103.

\bibitem[{Ryu et~al.(2017)Ryu, Kim, Choi, Yu, and Lee}]{Ryu2017ae}
Seonghan Ryu, Seokhwan Kim, Junhwi Choi, Hwanjo Yu, and Gary~Geunbae Lee. 2017.
\newblock Neural sentence embedding using only in-domain sentences for
  out-of-domain sentence detection in dialog systems.
\newblock \emph{Pattern Recogn. Lett.}, 88(C):26--32.

\bibitem[{Ryu et~al.(2018)Ryu, Koo, Yu, and Lee}]{ryu2018ood}
Seonghan Ryu, Sangjun Koo, Hwanjo Yu, and Gary~Geunbae Lee. 2018.
\newblock Out-of-domain detection based on generative adversarial network.
\newblock In \emph{EMNLP}, pages 714--718.

\bibitem[{Shen et~al.(2019{\natexlab{a}})Shen, Nama, and Jin}]{Shen2019teach}
Yilin Shen, Sandeep Nama, and Hongxia Jin. 2019{\natexlab{a}}.
\newblock Teach once and use everywhere -- building ai assistant eco-skills via
  user instruction and demonstration (poster).
\newblock In \emph{MobiSys}, pages 606--607.

\bibitem[{Shen et~al.(2019{\natexlab{b}})Shen, Ray, Jin, and
  Nama}]{shen2019skillbot}
Yilin Shen, Avik Ray, Hongxia Jin, and Sandeep Nama. 2019{\natexlab{b}}.
\newblock {S}kill{B}ot: Towards automatic skill development via user
  demonstration.
\newblock In \emph{NAACL-HLT, System Demonstrations}, pages 105--109.

\bibitem[{Shen et~al.(2018{\natexlab{a}})Shen, Ray, Patel, and
  Jin}]{shen2018cruise}
Yilin Shen, Avik Ray, Abhishek Patel, and Hongxia Jin. 2018{\natexlab{a}}.
\newblock {CRUISE:} cold-start new skill development via iterative utterance
  generation.
\newblock In \emph{ACL, System Demonstrations}, pages 105--110.

\bibitem[{Shen et~al.(2019{\natexlab{c}})Shen, Wang, Patel, and
  Jin}]{shen2019QA}
Yilin Shen, Yu~Wang, Abhishek Patel, and Hongxia Jin. 2019{\natexlab{c}}.
\newblock Sliqa-i: Towards cold-start development of end-to-end spoken language
  interface for question answering.
\newblock In \emph{ICASSP}, pages 7195--7199.

\bibitem[{Shen et~al.(2019{\natexlab{d}})Shen, Zeng, and Jin}]{shen2019slot}
Yilin Shen, Xiangyu Zeng, and Hongxia Jin. 2019{\natexlab{d}}.
\newblock A progressive model to enable continual learning for semantic slot
  filling.
\newblock In \emph{EMNLP-IJCNLP}, pages 1279--1284.

\bibitem[{Shen et~al.(2018{\natexlab{b}})Shen, Zeng, Wang, and
  Jin}]{Shen2018interspeech}
Yilin Shen, Xiangyu Zeng, Yu~Wang, and Hongxia Jin. 2018{\natexlab{b}}.
\newblock User information augmented semantic frame parsing using progressive
  neural networks.
\newblock In \emph{INTERSPEECH}, pages 3464--3468.

\bibitem[{Tan et~al.(2019)Tan, Yu, Wang, Wang, Potdar, Chang, and
  Yu}]{tan2019ood}
Ming Tan, Yang Yu, Haoyu Wang, Dakuo Wang, Saloni Potdar, Shiyu Chang, and
  Mo~Yu. 2019.
\newblock Out-of-domain detection for low-resource text classification tasks.
\newblock In \emph{EMNLP-IJCNLP}, pages 3564--3570.

\bibitem[{Tur and De~Mori(2011)}]{tur2011spoken}
G.~Tur and R.~De~Mori. 2011.
\newblock \emph{Spoken Language Understanding: Systems for Extracting Semantic
  Information from Speech}.
\newblock Wiley.

\bibitem[{Tur et~al.(2014)Tur, Deoras, and Hakkani-T{\"u}r}]{tur2014detecting}
Gokhan Tur, Anoop Deoras, and Dilek Hakkani-T{\"u}r. 2014.
\newblock Detecting out-of-domain utterances addressed to a virtual personal
  assistant.
\newblock In \emph{Fifteenth Annual Conference of the International Speech
  Communication Association}.

\bibitem[{Wang et~al.(2005)Wang, Deng, and Acero}]{WangDengAce:05}
Ye-Yi Wang, Li~Deng, and Alex Acero. 2005.
\newblock Spoken language understanding.
\newblock \emph{IEEE Signal Processing Magazine}, 22(5):16--31.

\bibitem[{Wang et~al.(2018)Wang, Shen, and Jin}]{Yu2018bimodel}
Yu~Wang, Yilin Shen, and Hongxia Jin. 2018.
\newblock A bi-model based rnn semantic frame parsing model for intent
  detection and slot filling.
\newblock In \emph{NAACL-HLT}, pages 309--314.

\bibitem[{Wolf et~al.(2019)Wolf, Debut, Sanh, Chaumond, Delangue, Moi, Cistac,
  Rault, Louf, Funtowicz, and Brew}]{Wolf2019HuggingFacesTS}
Thomas Wolf, Lysandre Debut, Victor Sanh, Julien Chaumond, Clement Delangue,
  Anthony Moi, Pierric Cistac, Tim Rault, R{\'{e}}mi Louf, Morgan Funtowicz,
  and Jamie Brew. 2019.
\newblock Huggingface's transformers: State-of-the-art natural language
  processing.
\newblock \emph{CoRR}, abs/1910.03771.

\bibitem[{Zhang and Wang(2016)}]{Zhang2016slu}
Xiaodong Zhang and Houfeng Wang. 2016.
\newblock A joint model of intent determination and slot filling for spoken
  language understanding.
\newblock In \emph{IJCAI}, pages 2993--2999.

\bibitem[{Zheng et~al.(2020)Zheng, Chen, and Huang}]{zheng2019ood}
Yinhe Zheng, Guanyi Chen, and Minlie Huang. 2020.
\newblock Out-of-domain detection for natural language understanding in dialog
  systems.
\newblock \emph{{IEEE} {ACM} Trans. Audio Speech Lang. Process.},
  28:1198--1209.

\end{thebibliography}



\end{document}